%% file: paper.tex
\documentclass{article}

\usepackage[accepted]{sysml2019}

\usepackage{microtype}
\usepackage{graphicx}
\usepackage{subfigure}
\usepackage{booktabs} 
\usepackage{tabularx}
\usepackage{upgreek}
\usepackage{comment}
\usepackage{amsmath,array,graphicx}

\usepackage{tabularx}
\usepackage{multirow}
\usepackage{graphicx}
\usepackage{algorithm}
\usepackage[colorinlistoftodos]{todonotes}
\usepackage{upgreek}

\usepackage[draft]{hyperref}

\usepackage{listings}
\usepackage[T1]{fontenc}
\usepackage[utf8]{inputenc}
\usepackage[scaled=0.8]{beramono}
\usepackage{amsmath}
\usepackage{amsfonts}
\usepackage{amssymb}
\usepackage{xcolor}
\usepackage{xspace}
\definecolor{light-gray}{gray}{0.45}
\definecolor{dark-green}{rgb}{0.0,0.34,0.25}

\newcommand{\norm}[1]{\left\lVert#1\right\rVert}
\newcommand{\algoName}[0]{\textsc{Condensa}\xspace}

\usepackage{eqparbox,array}

\newcommand{\argmin}{\operatornamewithlimits{argmin}}

\newcommand{\bx}{{\bf x}}
\newcommand{\bs}{{\bf s}}

\newcommand{\bw}{{\bf w}}

\sysmltitlerunning{A Programmable Approach to Neural Network Compression}

\begin{document}

\twocolumn[
\sysmltitle{A Programmable Approach to Neural Network Compression}




\begin{sysmlauthorlist}
\sysmlauthor{Vinu Joseph}{uou}
\sysmlauthor{Saurav Muralidharan}{nvr}
\sysmlauthor{Animesh Garg}{uot}
\sysmlauthor{Michael Garland}{nvr}
\sysmlauthor{Ganesh Gopalakrishnan}{uou}
\end{sysmlauthorlist}

\sysmlaffiliation{uou}{University of Utah}
\sysmlaffiliation{uot}{University of Toronto}
\sysmlaffiliation{nvr}{NVIDIA}

\sysmlcorrespondingauthor{Vinu Joseph}{vinu@cs.utah.edu}

\sysmlkeywords{Model Compression, Bayesian Optimization}

\vskip 0.3in

\begin{abstract}
Deep neural networks (DNNs) frequently contain far more weights, represented at a higher precision, than are required for the specific task which they are trained to perform. Consequently, they can often be compressed using techniques such as weight pruning and quantization that reduce both the model size and inference time without appreciable loss in accuracy. However, finding the best compression strategy and corresponding target sparsity for a given DNN, hardware platform, and optimization objective currently requires expensive, frequently manual, trial-and-error experimentation. In this paper, we introduce a programmable system for model compression called Condensa. Users programmatically compose simple operators, in Python, to build more complex and practically interesting compression strategies. Given a strategy and user-provided objective (such as minimization of running time), Condensa uses a novel Bayesian optimization-based algorithm to automatically infer desirable sparsities. Our experiments on four real-world DNNs demonstrate memory footprint and hardware runtime throughput improvements of 188x and 2.59x, respectively, using at most ten samples per search. We have released a reference implementation of Condensa at~\url{https://github.com/NVlabs/condensa}.
\end{abstract}
]



\printAffiliationsAndNotice{}  

\input{tex/introduction.tex}
\input{tex/related.tex}
\input{tex/condensa.tex}
\input{tex/results.tex}
\input{tex/conclusions.tex}

\bibliography{paper}
\bibliographystyle{sysml2019}



\end{document}

%% file: tex/introduction.tex
\section{Introduction}
Modern deep neural networks (DNNs) are complex,
and often contain millions of parameters spanning
dozens or even hundreds of layers~\cite{he2016deep,huang:2017}.
This complexity translates into substantial memory and runtime costs
on hardware platforms at all scales.
Recent work has demonstrated that DNNs are often over-provisioned and can be compressed without appreciable loss of accuracy.
Model compression can be used to
reduce both model memory footprint and inference latency using techniques such as
weight pruning~\cite{han2015learning,luo2017thinet},
quantization~\cite{gupta2015deep}, and low-rank
factorization~\cite{jaderberg2014speeding,denton2014exploiting}.
Unfortunately, the requirements of
different {\em compression contexts}---DNN structure,
target hardware platform, and the user's optimization objective---are often in conflict.
The recommended compression strategy for reducing inference latency
may be different from that required to reduce total memory footprint.
For example, in a Convolutional Neural Network (CNN),
reducing inference latency may require pruning filters to realize speedups on real hardware~\cite{li2016pruning}, while reducing memory footprint may be accomplished by zeroing out individual weights.
Similarly, even for the {\em same optimization objective},
say reducing inference latency, one may employ filter pruning for a CNN,
while pruning 2-D blocks of non-zero weights~\cite{gray:2017} for a
language modeling network such as Transformer~\cite{vaswani:2017},
since the latter has no convolutional layers.
Thus, it is crucial to enable convenient expression of 
alternative compression schemes, yet
none of today's model compression approaches help the designer
tailor compression schemes to their needs.

\begin{figure}[tb]
\centering
\includegraphics[width=\linewidth]{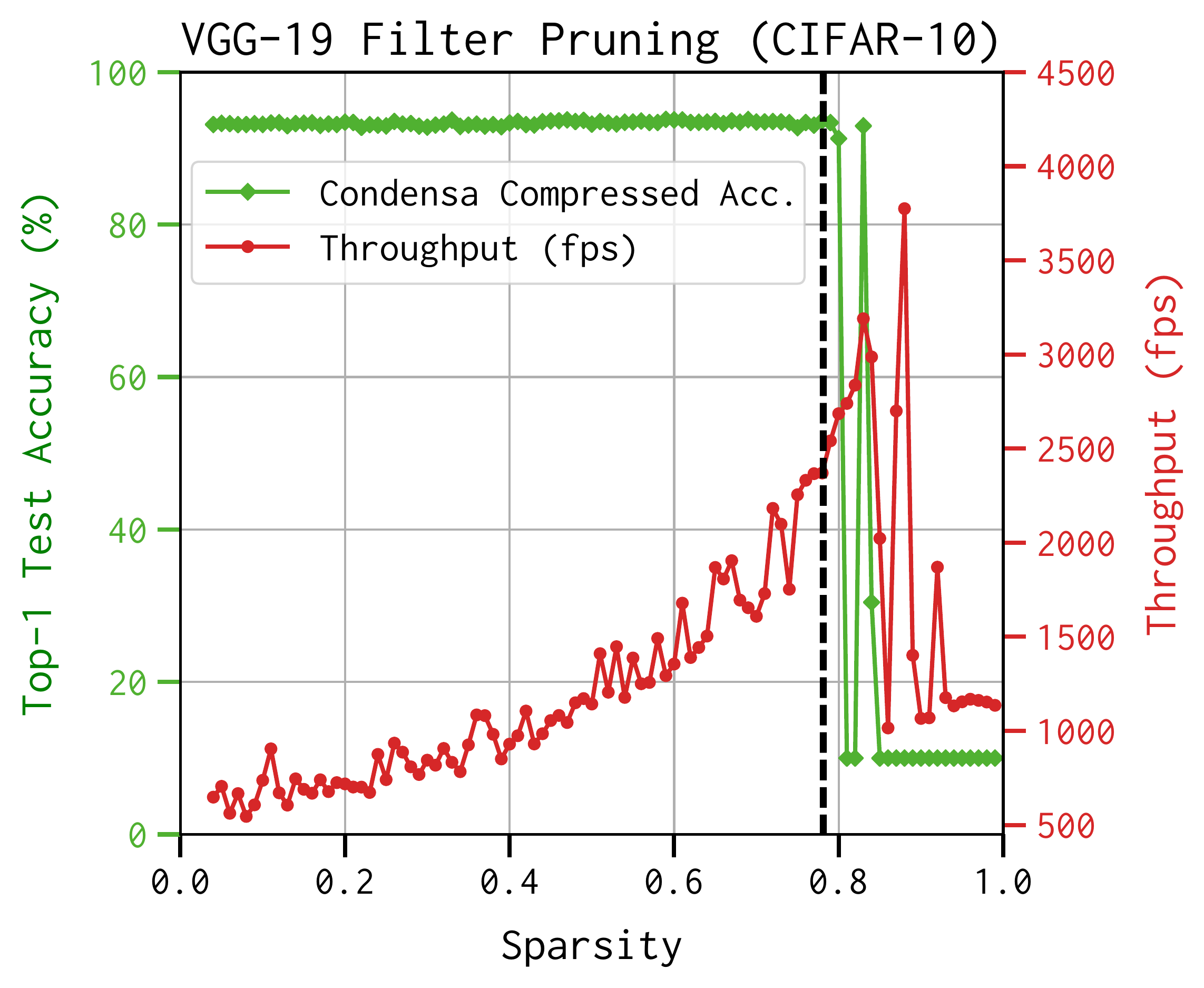}
\caption{Top-1 test accuracy (green) and throughput (red) vs.\ sparsity for VGG-19 on CIFAR-10.
\algoName is designed to solve constrained optimization problems of the form ``maximize throughput, with a lower bound on accuracy". In this case, \algoName automatically discovers a sparsity (vertical dashed line) and compresses the model to this sparsity,
improving throughput by $2.59\times$ and accuracy by $0.36\%$.
}
\vspace{-10pt}
\label{fig:vgg-intro}
\end{figure}

%
Current approaches to model compression
also require manual specification of compression hyperparameters, such
as {\bf target sparsity}---{\em the proportion of zero-valued parameters in the
compressed model vs.\ the original.}
However, with current approaches, finding the best sparsity 
often becomes a trial-and-error search, with
each such trial having a huge cost (often multiple days for large models such as BERT) and involving training the compressed model to convergence,
only to find (in most cases) that the compression objectives are not met.
The main difficulty faced by such unguided approaches is
that sparsities 
vary unpredictably with changes in the compression context,
making it very difficult to provide users with any guidelines, whatsoever.
Therefore, automatic and {\em sample-efficient} approaches that minimize the number of trials are crucial
to support the needs of designers who must adapt
a variety of neural networks to a broad spectrum of platforms targeting a wide
range of tasks.

To address the above-mentioned problems of flexible expression of compression strategies, automated compression hyperparameter inference, and sample efficiency, we introduce \algoName, a new framework for programmable model compression. As an illustration of the level of automation provided by \algoName,
consider the problem of improving the
inference throughput of VGG-19~\cite{simonyan2014very} on the CIFAR-10 image
classification task~\cite{krizhevsky2014cifar}.
Since VGG-19 is a convolutional neural network, one way to improve its
inference performance on modern hardware such as GPUs is by pruning
away individual convolutional filters~\cite{he2018progressive}.
Figure~\ref{fig:vgg-intro} shows the accuracy and throughput obtained
by \algoName on this task.
Here, we plot the compressed model's top-1 test accuracy and throughput as a function of the sparsity (green and red lines,
respectively).\footnote{Note that these curves are not known a priori and
are often extremely expensive to sample;
they are only plotted here to better place the obtained solution in context.}
\algoName's solution corresponds to a sparsity of $0.79$
and is depicted as the vertical dashed line.
This result is significant for two reasons: (1) using the \algoName library,
the filter pruning strategy employed for this experiment was expressed in
less than 10 lines of Python code, and (2) the optimal sparsity of
$0.79$ that
achieves throughput and top-1 accuracy improvements of $2.59\times$ and $0.36\%$, respectively,
was obtained automatically by \algoName using a sample-efficient constrained
Bayesian optimization algorithm.
Here, the user didn't have to specify any
sparsities manually, and instead only had to define a domain-specific
objective function to maximize (inference throughput, in this case).

This paper makes the following contributions:
\begin{enumerate}
    \item It presents \algoName, a new framework for programmable neural network compression. \algoName supports the expression of
the overall compression strategy in Python using operators provided by its compression library. 
Since each strategy is a Python function, users are  
able to programmatically compose elementary schemes to build much
more complex and practically interesting schemes.

\item It presents a novel sample-efficient algorithm based on Bayesian optimization (B.O.) in \algoName for automatically inferring optimal sparsities based on a user-provided objective function. Given \algoName's ability to support the expression of meaningful high-level
objective functions---for example, the throughput (images/sec) of a convolutional neural network---users
are freed from the burden of having to specify compression hyperparameters manually.

\item It demonstrates the effectiveness of \algoName on three image classification and language modeling tasks, resulting in memory footprint reductions of up to $188\times$ and runtime throughput improvements of up to $2.59\times$ using at most 10 samples per search.
\end{enumerate}

%% file: tex/related.tex
\section{Background}

For a given task such as image classification, assume we have trained a large {\em reference} model $\overline{\bw} = \argmin_{\bw} L(\bw)$, where $L()$ denotes a {\em loss function} (e.g., cross-entropy on a given training set), and $\bw \in \mathbb{R}^P$. {\em Model compression} refers to finding a smaller model $\Theta$ that can be applied to the same task and ideally achieves the same accuracy as $\overline{\bw}$.
Model compression can be performed in various ways, and \algoName currently supports two commonly used techniques: pruning and quantization. In pruning, non-zero values from $\overline{\bw}$ are eliminated or ``pruned'' to obtain $\Theta$. Pruning is usually performed using some kind of thresholding (for eg., magnitude-based) and can be unstructured (prune any non-zero value) or structured (prune only {\em blocks} of non-zeros). On the other hand, quantization retains the number of parameters in $\Theta$ but assigns parameters in $\overline{\bw}$ one of K codebook values, where the codebook may be fixed or adaptive. \algoName supports low-precision approximation, which refers to assigning each parameter in $\overline{\bw}$ a corresponding lower-precision representation (for example, converting from 32-bit to 16-bit floating-point) and is equivalent to quantization using a fixed codebook.

\noindent \textbf{DNN Compression Techniques}
There is considerable prior work on accelerating neural networks using structured 
weight pruning~\cite{wang2019structured,mccarley2020structured,frankle2018lottery, han2015learning, luo2017thinet, han2017ese, dong2017more, han2016eie, polyak2015channel, hu2016network, anwar2016compact, molchanov2016pruning}, quantization~\cite{zhu2016trained, gong2014compressing} and 
low-rank tensor factorization~\cite{kossaifi2020factorized,lebedev2014speeding, xue2013restructuring, denton2014exploiting, girshick2015fast}.
Most of these individual compression
schemes for pruning and quantization and their combinations can be expressed in \algoName. Two common problems with these existing methods are: (1) determining optimal sparsity at a global (network) level, and (2) distributing global sparsity into per-layer sparsities.
We tackle these problems efficiently and systematically using our Bayesian and L-C optimizers, respectively, as described in Section~\ref{sec:condensa}.

\noindent \textbf{Automated Model Compression}
Automating model compression involves finding both an optimal compression strategy for a given $\overline{\bw}$, along with its corresponding compression hyperparameters such as target sparsity with minimal manual intervention. Current state-of-the-art frameworks in this domain include AMC~\cite{he2018amc} and AutoCompress~\cite{liu2019autoslim}, which use reinforcement learning and simulated annealing, respectively, to automatically find desirable target sparsities for a fixed pruning strategy. \algoName, in contrast, supports the programmable expression of a wide variety of compression strategies (not just pruning). Also, in the context of automated model compression, each sample corresponds to training the compressed model to convergence, and can be extremely expensive to compute; unfortunately, techniques such as reinforcement learning, which is used in AMC~\cite{he2018amc}, can be highly sample-inefficient~\cite{mnih2013playing}. To minimize the number of samples drawn, \algoName uses a novel and sample-efficient Bayesian optimization-based algorithm for automatically arriving at desirable target sparsities. While Bayesian optimization has previously been demonstrated to work well for general hyperparameter optimization in machine learning and neural architecture search~\cite{snoek2012practical,dai2019chamnet}, to the best of our knowledge, we are the first to use sample-efficient search via Bayesian optimization for obtaining compression hyperparameters.

\noindent \textbf{General Compression Algorithms and Tools} General accuracy recovery algorithms capable of handling a wide variety of compression techniques provide the foundation for systems like \algoName. Apart from the L-C algorithm~\cite{carreira2017model} which \algoName uses, other recent accuracy recovery algorithms have been proposed. ADAM-ADMM~\cite{zhang2018adam} proposes a unified framework for structured weight pruning based on ADMM that performs dynamic regularization in which the regularization target is updated in each iteration. DCP~\cite{zhuang2018discrimination} introduces additional losses into the network to increase the discriminative power of intermediate layers and select the most discriminative channels for each layer by considering the additional loss and the reconstruction error. \algoName can readily support such algorithms as additional optimizers as described in Section~\ref{sec:condensa}. Neural network distiller~\cite{neta_zmora_2018_1297430}, TensorFlow model optimization toolkit~\cite{tftoolkit} {\color{black} and NNCF~\cite{kozlov2020neural} are three recent open-source model compression frameworks that support multiple compression schemes.} While these projects share a number of common goals with \algoName, they differ in two important ways: first, they do not support the expression of schemes as imperative programs containing control-flow, iteration, recursion, etc.~(Distiller requires a declarative compression specification in YAML, while the TensorFlow model optimization toolkit operates by modifying the DNN computation graph directly); second, these frameworks do not support automatic compression hyperparameter optimization for black-box objective functions.

%% file: tex/condensa.tex
\section{Condensa Framework}
\label{sec:condensa}


\begin{figure*}[tbp]
\centering
\includegraphics[width=\textwidth]{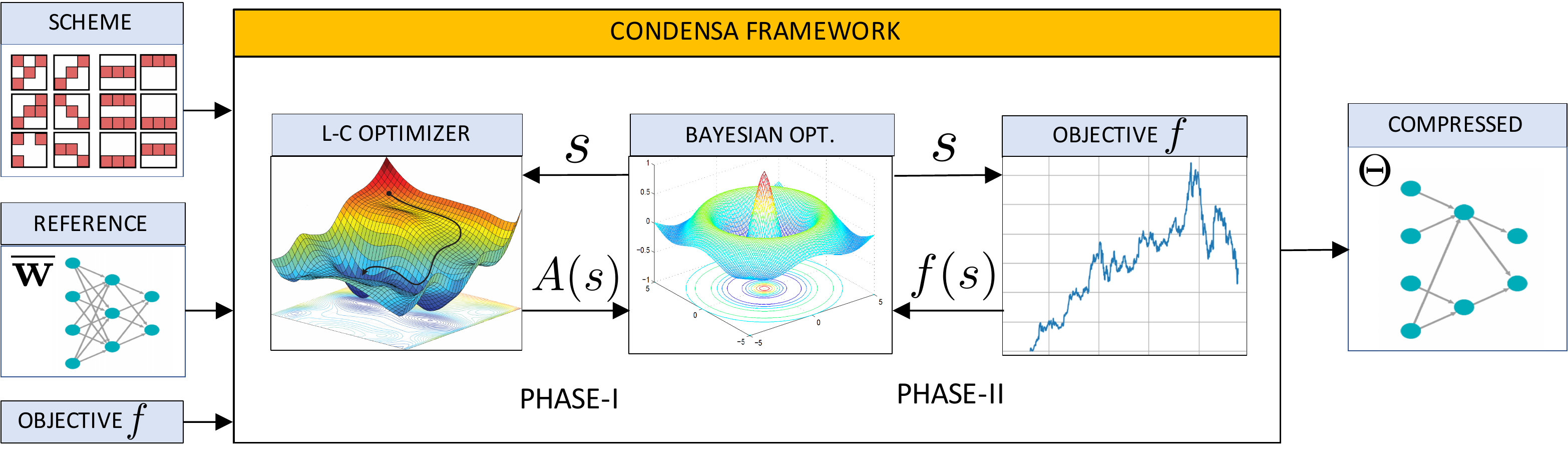}
\caption{\algoName framework overview. The user provides the pre-trained model ($\overline{\bw}$), a compression scheme, and an objective function $f$. \algoName uses the Bayesian and L-C optimizers to infer an optimal target sparsity $s^*$ and corresponding compressed model $\Theta$.}
\label{fig:condensa}  
\end{figure*}

Figure~\ref{fig:condensa} provides a high-level overview of the \algoName framework. As shown on the left-hand side of the figure, a user compresses a pre-trained model $\overline{\bw}$ by specifying a compression scheme and an objective function $f$. Both the scheme and objective are specified in Python using operators from the \algoName library; alternatively, users may choose from a selection of commonly used built-in schemes and objectives. The \algoName library is described in more detail in Section~\ref{sec:library}. Apart from the operator library, the core framework, shown in the middle of the figure, consists primarily of two components: (1) the constrained Bayesian optimizer for inferring optimal sparsities, and (2) the L-C optimizer for accuracy recovery. These components interact with each other as follows: at each iteration, the Bayesian optimizer samples a sparsity $s$, which is fed into the L-C optimizer. The L-C optimizer distributes this global sparsity across all the layers of the network and performs accuracy recovery (this process is described in more detail in Section~\ref{sec:lc}), passing the final obtained accuracy $A(s)$ back to the Bayesian optimizer. The compressed model $w$ obtained by the L-C optimizer is also used to evaluate the user-provided objective function $f$, the result of which is fed into the Bayesian optimizer. Based on these inputs ($A(s)$ and $f(w)$), the Bayesian optimizer decides the next point to sample. The sparsity that satisfies both the accuracy and objective constraints ($s^*$) is used to obtain the final compressed model (denoted by $\Theta$ in the figure). The Bayesian and L-C optimizers are described in more detail in Sections~\ref{sec:bo} and~\ref{sec:lc}, respectively.

\subsection{Condensa Library}
\label{sec:library}

The \algoName Library provides a set of operators for constructing complex compression schemes programmatically in Python. Three sets of operators are currently supported: (1) the \texttt{quantize} and \texttt{dequantize} operators for converting network parameters from a 32-bit floating-point representation to a lower-precision one such as 16-bit floating-point, and in the opposite direction, respectively; (2) the \texttt{prune} operator for unstructured magnitude-based pruning, and (3) the \texttt{filter\_prune}, \texttt{neuron\_prune}, and \texttt{blockprune} operators for pruning blocks of nonzeros (structure pruning). Each operator can be applied on a per-layer basis.

\algoName's tight integration with the Python ecosystem makes the expression of common compression patterns more natural. For example, operators can be combined with conditional statements to selectively compress layers based on properties of the input DNN and/or target hardware platform, as shown below:

\begin{lstlisting}[numbers=none]
# Prune only non-projection layers in ResNets
if not layer.is_projection: prune(layer)
# Quantize only if FP16 hardware is available
if platform_has_fast_fp16(): quantize(layer)
\end{lstlisting}

Similarly, the use of iteration statements obviates the need for applying compression operators individually for each layer, resulting in more concise and readable schemes. This is in contrast to frameworks such as Distiller~\cite{neta_zmora_2018_1297430} which require a per-layer declarative compression specification.

\paragraph{Pre-built Schemes} In addition to the layer-wise operators described above, the \algoName Library also includes a set of pre-built compression {\em schemes} that operate on the full model. \algoName includes schemes for unstructured and structured pruning, quantization, and composition of individual schemes. These schemes handle a number of low-level details such as magnitude threshold computation from a sparsity, filter/neuron/block aggregation, etc., enabling non-expert users to quickly get started with \algoName without knowledge of low-level implementation details. The current set of pre-built schemes is listed in Table~\ref{tab:schemes}.

Listing~\ref{code:condensa} provides a concrete example of invoking \algoName to compress a model. Here, we first train the reference models
(lines 2-3) and instantiate the pre-built \texttt{FilterPrune} scheme for structured pruning (line 6). We also define our objective function to be throughput (line 8) and specify that it must be maximized (line 10); note that while users may define their own objective functions, \algoName also comes bundled with some common objective functions such as model memory footprint and throughput. Next, we instantiate the L-C optimizer (line 12) and the model compressor (lines 14-24). The model compressor (\texttt{Compressor} class in Listing) automatically samples and evaluates global sparsities as described in Section~\ref{sec:bo} and returns the final compressed model.

\begin{figure}[tb]
\begin{lstlisting}[label=code:condensa,caption={Example usage of the \algoName library.}]
# Construct pre-trained model
criterion = torch.nn.CrossEntropyLoss()
train(model, num_epochs, trainloader, criterion)

# Instantiate compression scheme
prune = condensa.schemes.FilterPrune()
# Define objective function
tput = condensa.objectives.throughput
# Specify optimization operator
obj = condensa.searchops.Maximize(tput)
# Instantiate L-C optimizer
lc = condensa.optimizers.LC(steps=30, lr=0.01)
# Build model compressor instance
compressor = condensa.Compressor(
    model=model, # Trained model
    objective=obj, # Objective
    eps=0.02, # Accuracy threshold
    optimizer=lc, # Accuracy recovery
    scheme=prune, # Compression scheme
    trainloader=trainloader, # Train dataloader
    testloader=testloader, # Test dataloader
    valloader=valloader, # Val dataloader
    criterion=criterion # Loss criterion
)
# Obtain compressed model
wc = compressor.run()
\end{lstlisting}
\end{figure}

\begin{table}[tb]
\centering
        \begin{tabularx}{\linewidth}{ l  X }
				\hline
				\textbf{Scheme} & \textbf{Description} \\ \hline
				\texttt{Quantize(dtype)} & Quantizes network weights to given datatype \texttt{dtype}. \\ \hline
				\texttt{Prune()} & Performs unstructured pruning of network weights. \\ \hline
                \texttt{NeuronPrune(criteria)} & Aggregates and prunes neurons (1D blocks) according to \texttt{criteria}.\\ \hline
				\texttt{FilterPrune(criteria)} & Aggregates and prunes filters (3D blocks) according to \texttt{criteria}. \\ \hline
				\texttt{StructurePrune(criteria)} & Combines neuron and filter pruning.\\ \hline
				\texttt{BlockPrune(criteria, bs)} & Aggregates and prunes n-D blocks of size \texttt{bs} according to \texttt{criteria}.\\ \hline
				\texttt{Compose(slist)} & Composes together all schemes in \texttt{slist}.\\ \hline
           \end{tabularx}
	\caption{List of pre-built compression schemes in \algoName.}
	\label{tab:schemes}
\end{table}

\input{tex/condensa-bo.tex}

\subsection{Accuracy Recovery using L-C}
\label{sec:lc}

As described earlier in this section, given a reference model, compression scheme, and compression hyperparameter values (obtained automatically by the Bayesian hyperparameter optimization subsystem described in Section~\ref{sec:bo}), \algoName tries to recover any accuracy lost due to compression.
While the compressed model, denoted as $\Theta$, can be obtained by directly zeroing out lower-magnitude parameters from the reference model $\overline{\bw}$ (a technique referred to as {\em direct compression}), the resulting model $\Theta$ is generally sub-optimal w.r.t. the loss since the latter is ignored in learning $\Theta$. Instead, we desire an {\em accuracy recovery algorithm} that obtains an {\em optimally compressed model} with locally optimal loss.
An effective accuracy recovery mechanism for \algoName must ideally have three important attributes: (1) able to handle all the compression operators supported by \algoName, (2) be efficient with relatively low overheads, and (3) provide optimality guarantees whenever possible.
In this paper, we use the recently proposed L-C algorithm~\cite{carreira2018learning}, since it satisfies all three of the above requirements. In L-C, model compression is formulated as a constrained optimization problem:
\begin{equation}
\text{min}_{\bw, \bf{\Theta}} L(\bw) \hspace{5 mm} \text{s.t.} \hspace{5 mm} \bw = \mathcal{D}(\Theta)
\label{eq:opt}
\end{equation}
Here, the \emph{decompression mapping} $\mathcal{D} :\Theta \in \mathbb{R}^{Q} \rightarrow \bw \in \mathbb{R}^{P}$ maps a low-dimensional parameterization to uncompressed model weights, and the \emph{compression mapping} $\mathcal{C}(\bw) = \argmin_{\Theta} \norm{\bw - \mathcal{D}(\Theta)}^{2}$ behaves
similar to the inverse of $\mathcal{D}$.
This formulation naturally supports a number of well-known compression techniques. In particular, pruning is defined as $\bw = \mathcal{D}(\Theta) = \Theta$ where $\bw$ is real and $\Theta$ is constrained to have fewer nonzero values by removing (zeroing out) lower magnitude weights; low-precision approximation defines a constraint $w_i = \theta_i$ per parameter where $w_i$ is in a higher-precision representation and $\theta_i$ is in a lower-precision one.

Eq.~\ref{eq:opt} is non-convex due to two reasons: (1) the original problem of training the reference model is already non-convex for models such as DNNs, making the objective function of Eq~\ref{eq:opt} non-convex, and (2) the decompression mapping $\mathcal{D}(\Theta)$ typically adds another layer of non-convexity caused by an underlying combinatorial problem. While a number of non-convex algorithms may be used to solve Eq~\ref{eq:opt}, we focus on the augmented Lagrangian (AL) method~\cite{wright1999numerical} implemented in the L-C algorithm~\cite{carreira2018learning} in this paper, since it is relatively efficient and easy to implement. As its name indicates, the L-C algorithm alternates between two steps: a learning (L) step which trains the uncompressed model but with a quadratic regularization term, and a compression (C) step, which finds the best compression of $\bw$ (the current uncompressed model) and corresponds to the definition of compression mapping $\mathcal{C}$. We refer the reader to~\cite{carreira2018learning} for a more detailed description of the L-C algorithm.
Other recent AL-based algorithms that could potentially be used include ADMM~\cite{zhang2018adam} and DCP~\cite{zhuang2018discrimination}.

\subsection{Implementation}
\label{sec:impl}

The \algoName library and L-C optimizer are implemented in Python and are designed to interoperate seamlessly with the PyTorch framework~\cite{pytorch}. While we chose PyTorch for its widespread use in the machine learning community, it is worth noting that \algoName's design is general and that its features can be implemented in other similar frameworks such as TensorFlow~\cite{abadi2016tensorflow} and MXNET~\cite{chen2015mxnet}. We currently use a publicly available Python library for Bayesian global optimization with Gaussian Processes~\cite{fmfnbo}.

\noindent \textbf{Network Thinning} Condensa comes pre-built with three {\em structure pruning} schemes: filter, neuron, and block pruning, as shown in Table~\ref{tab:schemes}. The application of these schemes may yield {\em zero structures}, which refer to blocks of zeros within a DNN's parameters. {\em Network thinning} refers to the process of identifying and removing such zero structures and consequently reducing the number of floating-point operations executed by the target hardware platform. Condensa employs a three-phase network thinning algorithm for structure pruning: in the first phase, we construct an in-memory graph representation of the target DNN. PyTorch makes this non-trivial, as its eager execution semantics preclude it from ever building a full graph-based representation of the DNN. To overcome this, we trace a forward execution path of the DNN and use it to construct an in-memory representation based on the ONNX format. In the next phase, we create a {\em thinning strategy} by analyzing the dependencies between the nodes of the graph constructed in the first phase. This step primarily involves keeping track of tensor dimension changes in a node due to thinning and ensuring that the corresponding tensor dimensions of the node's successors are appropriately adjusted. Due to the possibility of complex dependence patterns such as skip nodes in real-world DNNs (for example, deep residual networks~\cite{he2016deep}), this step is the most challenging to implement. In the final phase, we apply the thinning strategy obtained in phase 2 and physically alter tensor shapes to obtain the final thinned network.
The Condensa Library provides the \texttt{thin} method which can be used to thin a given compressed model.

%% file: tex/condensa-bo.tex

\subsection{Sample-Efficient Bayesian Optimization}
\label{sec:bo}

It is intuitive to split the problem of finding optimal target sparsities into two stages: (1) find the highest target sparsity that loses at most $\epsilon$ accuracy w.r.t the original uncompressed model $\overline{\bw}$, and (2) in a constrained sparsity regime obtained from stage (1), optimize a user-provided objective function $f$ (e.g., throughput, or memory footprint) and return the solution as the final sparsity.
For both stages, \algoName utilizes Bayesian optimization as shown in Figure~\ref{fig:condensa}.

Bayesian Optimization (B.O.) is an optimization framework based on continually updating a {\em probabilistic model} with measurements of a function to be optimized~\cite{jones1998efficient}. Given a set of parameters to be optimized, B.O. makes black-box calls to the objective, updates the probabilistic model with the new information, and selects the next point to evaluate using an {\em acquisition function} that combines information about the expectation and uncertainty of a function value under the probabilistic model.
\algoName employs a Gaussian Process (G.P.) model for B.O. due to its favorable statistical and computational characteristics~\cite{srinivas2009gaussian}. 
It is worth highlighting that B.O. leverages principled Bayesian inference to trade off exploration and exploitation, and is sample-efficient for non-convex black-box functions such as the ones optimized by \algoName
\cite{jones1998efficient}.

In \algoName's two-stage optimization pipeline, we first find a sparsity $s_{acc}$ that constrains the model accuracy function $A$ to the provided $\epsilon$. We then constrain the {\em sparsity search space} to $(0, s_{acc})$ while optimizing the user-provided objective function $f$. Note that we assume that $A$ decreases monotonically w.r.t. sparsity in the region $(0, s_{acc})$.
For each stage, \algoName uses a distinct acquisition function
to guide the next best point for function evaluation.

\paragraph{Stage 1: Solving Accuracy Constraints}
Recall that in the first stage of the sparsity inference process, we aim to find the highest sparsity $s_{acc}$ that loses at most $\epsilon$ accuracy w.r.t. the original reference model $\overline{\bw}$. To this end, we first define a {\em Level-Set} $L$ that represents $Acc(\overline{\bw}) - \epsilon$ and aim to find the point on the accuracy curve of the compressed model that intersects with $L$; the sparsity corresponding to this solution will be $s_{acc}$. We propose a novel acquisition function to find $s_{acc}$ named Domain-Restricted Upper Confidence Bound (DR-UCB).

DR-UCB builds upon an existing level-set black-box optimization technique named ILS-UCB~\cite{garg2016tumor}, 
which is characterized by two properties: (1) it prioritizes searching in the region where the level set intersects the accuracy curve, (2) it does not seek to precisely learn the shape of the entire accuracy curve. However, in \algoName, since accuracy values can be safely assumed to decrease monotonically with increasing sparsity, we notice that it is also possible to progressively restrict the search domain of sparsities based on whether the currently sampled point meets the level-set constraints. In DR-UCB, we exploit this property to greatly improve sample efficiency over ILS-UCB. Mathematically, we define $\bs_t$, the sparsity value sampled at iteration $t$ using DR-UCB, as follows:
\begin{multline}
\bs_t = \underset{\bs_{}}{\text{argmax}}~(1 - \gamma) \sigma(\bs) - \gamma | \mu(\bs) - L| \\
    \text{s.t.} \quad \bs_t > \bs_{i} \quad \forall i \in [0, t-1], \quad \mathcal{B}_f(\bs_t)  \ge  L
    \label{eq:dr-ucb}
\end{multline}
Here, $\mathcal{B}_f$ represents the L-C accuracy function, and $\bs_t$ is (1) greater than all the previous sparsities $\bs_i$, and (2) satisfies the level set constraint $\mathcal{B}_f(\bs_t)  \ge  L$. We achieve this by minimizing the difference between the GP's mean curve $\mu(\bs)$ and the level set using the term $|\mu(\bs) - L|$ in (\ref{eq:dr-ucb});
the parameter $\gamma$ controls the trade-off between exploitation and exploration.
%
%
Algorithm~\ref{alg:bo} illustrates how DR-UCB is employed to efficiently find $s_{acc}$.

\newcommand\CONDITION[2]%
  {\begin{tabular}[t]{@{}l@{}l@{}}
     #1&#2
   \end{tabular}%
  }
  
  \algdef{SE}[IF]{If}{EndIf}[1]%
  {\algorithmicif\ \CONDITION{#1}{\ \algorithmicthen}}%
  {\algorithmicend\ \algorithmicif}%
\algdef{C}[IF]{IF}{ElsIf}[1]%
  {\algorithmicelse\ \algorithmicif\ \CONDITION{#1}{\ \algorithmicthen}}

\makeatletter
\algnewcommand{\LineComment}[1]{\Statex \hskip\ALG@thistlm \(\triangleright\) #1}
\makeatother

\begin{algorithm}[tb]
  \caption{Bayesian Sparsity Inference with Domain Restriction}
  \label{alg:bo}
  \begin{minipage}{.95\columnwidth}
    \begin{algorithmic}[1]
     \Procedure{BO$_{DR-UCB}$}{$\mathcal{B}_f$, $L$, $T$}
        \LineComment{$\mathcal{B}_f$: Function to optimize}
        \LineComment{$L$: Level set}
        \LineComment{$T$: \# Iterations}
        \State \texttt{GP} $\leftarrow$    \texttt{GP-Regressor.initialize()}
        \State $s_0$ $\leftarrow$ $0$; $D$ $\leftarrow$ $(0, 1)$; $\mathbf{X}$ $\leftarrow$ $\emptyset$
        \For{$t \gets 1, 2, \ldots$ . $T-1$}
          \State{$s_t \gets$ $\texttt{argmax}_{D}\text{DR-UCB}(s|\mathbf{X}_{0:t-1})$}
          \State{$y_t \gets \mathcal{B}_f(s_t)$}
          \If{$s_t > s_{t-1}$ \textbf{and} $y_t \ge$ $L$}
              \State{$D$ $\leftarrow$ $(s_t,1)$}
          
          \EndIf
        \State{$\mathbf{X}_{0:t} \gets \{\mathbf{X}_{0:t-1}, (s_t, y_t)\}$}
        \State{\texttt{GP.Update}($\mathbf{X}_{0:t}$)}
        \EndFor
        \State{\Return $s_{T-1}$}
      \EndProcedure
    \end{algorithmic}
  \end{minipage}
\end{algorithm}

\paragraph{Stage 2: Optimizing the User-Defined Objective} Once we find a  sparsity $s_{acc}$ that satisfies the user-provided accuracy constraints in stage 1, our next objective is to find the final sparsity $s^*$ that optimizes the user-defined objective function $f$ in the constrained sparsity domain $(0, s_{acc})$. For this, we employ the Upper and Lower Confidence Bound (UCB/LCB) acquisition functions for function maximization and minimization, respectively~\cite{srinivas2009gaussian}.

%% file: tex/results.tex
\section{Evaluation}
\label{sec:evaluation}

We conduct extensive experiments and fully analyze \algoName on three real-world tasks:


\noindent \textbf{(1) Image Classification on CIFAR-10}
The CIFAR-10 dataset~\cite{krizhevsky2014cifar} consists of $50k$ training and $10k$ testing $32 \times 32$ images in 10 classes. We train the VGG-19~\cite{simonyan2014very} and ResNet56~\cite{he2016deep} neural networks on this dataset for $160$ epochs with batch normalization, weight decay ($10^{-4}$), decreasing learning rate schedules (starting from $0.1$) and augmented training data.

\noindent \textbf{(2) Image Classification on ImageNet}
Here, we use the VGG-16 neural network~\cite{simonyan2014very} trained on the challenging ImageNet task~\cite{deng2009imagenet}, specifically the ILSVRC2012 version. We use PyTorch~\cite{pytorch} and default pretrained models as a starting point.

\begin{table*}[tbh]
  \centering
  \caption{%
\algoName performance results on CIFAR-10, ImageNet, and WikiText-2. Here, $s^{*}$ represents the target sparsity obtained by \algoName, $r_c$ is the memory footprint reduction, and $s_F$ the FLOP reduction. The level-set, represented by $\epsilon$, is set to $2\%$ below baseline in all experiments.
    }
\begin{center}
\begin{small}
\begin{sc}
  \label{tab:results}
 {\small
  \resizebox{\textwidth}{!}{
\centering
\scriptsize
\begin{tabular}{ccccrrrr}
\toprule
 Method &  Dataset & Network & $s^{*}$ & \multicolumn{1}{c}{Accuracy} & $r_c$ & Throughput \\
 \midrule
 \textbf{Baseline} & CIFAR-10 & VGG19-BN & & $92.98\%$ & 1$\times$ & 1$\times$\\
 \textbf{\algoName} P+Q & CIFAR-10  & VGG19-BN & $0.99$ & $93.26\%$ & $188.23\times$ & N/A \\
 \textbf{\algoName} Filter & CIFAR-10  & VGG19-BN & $0.79$ & $93.34\%$ & $1.35\times$ & $2.59\times$\\
 \midrule
 \textbf{Baseline} & CIFAR-10 & ResNet56 & & $92.75\%$ & 1$\times$ & 1$\times$\\
AMC~\cite{he2018amc} &  CIFAR-10  & ResNet56 & N/A & $90.1\%$ & N/A & $s_F=2\times$\\
 \textbf{\algoName} P+Q &  CIFAR-10  & ResNet56 & $0.95$ & $91.42\%$ & $31.14\times$ & N/A \\
 \textbf{\algoName} Filter & CIFAR-10  & ResNet56 & $0.63$ & $93.18\%$ & $1.14\times$ & $1.17\times$ \\
 \midrule
 \textbf{Baseline} & ImageNet & VGG16-BN & & $91.50\%$ & 1$\times$ & 1$\times$\\
 Filter Pruning~\cite{he2017channel} & ImageNet & VGG16-BN & & $89.80\%$ & $\approx4\times$ & N/A \\ 
 AutoCompress~\cite{liu2019autoslim} &  ImageNet & VGG16-BN & N/A & $90.90\%$ & $6.4\times$ & N/A \\
 AMC~\cite{he2018amc} & ImageNet & VGG16-BN & N/A & $90.1\%$ & N/A & $s_F=1.25\times$ \\
 \textbf{\algoName} P+Q & ImageNet & VGG16-BN & $0.93$ & $89.89\%$ & $29.29$ & N/A\\
 \textbf{\algoName} Filter & ImageNet & VGG16-BN & $0.12$ & $90.25\%$ & $1\times$ & $1.16\times$\\
 \midrule
 \textbf{Baseline} & WikiText-2 & LSTM & & Log-Perplexity: $4.70$ & 1$\times$ & 1$\times$\\
 Lottery Ticket\cite{yu2019playing} & WikiText-2 & LSTM & N/A & Log-Perplexity: $4.70$ & $\approx 10\times$ & N/A\\
 \textbf{\algoName} P+Q & WikiText-2 & LSTM & $0.92$ & Log-Perplexity: $4.75$ & $4.2\times$ & N/A \\
 \textbf{\algoName} Block & WikiText-2 & LSTM & $0.60$ & Log-Perplexity: $4.62$ & $1.1\times$ & $s_F=2.14$\\
 \bottomrule
 \end{tabular}
 }}
\end{sc}
\end{small}
\end{center}
\end{table*}

\begin{figure*}[tbh]
\centering
\includegraphics[width=0.92\linewidth, height=5.1in] {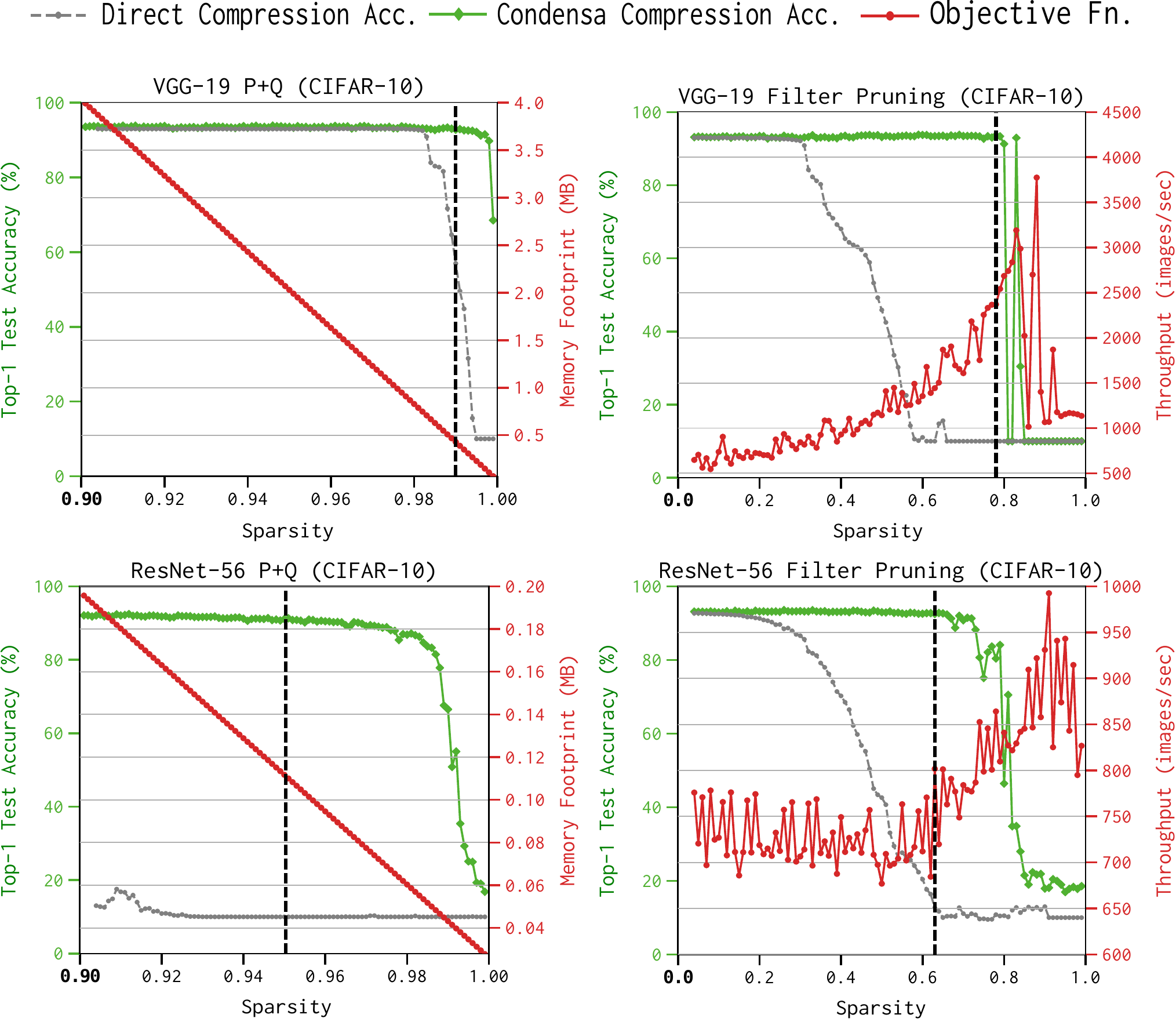}
\caption{\algoName sparsity profiles for VGG19-BN and ResNet56 for CIFAR-10. Column 1 shows the problem of the form ``minimize \textit{memory footprint} with a lower bound on accuracy", while Column 2 illustrates ``maximize \textit{throughput} with a lower bound on accuracy". The DC line (gray) shows accuracy values if no accuracy recovery with L-C is performed. Note that the x-axis ranges are different: the plots on the left have sparsities ranging from $0.9$ to $1.0$ while those on the right have values ranging from $0$ to $1$.}
\label{fig:profile-cifar}
\end{figure*}

\begin{figure*}
\centering
\includegraphics[height=0.4\linewidth] {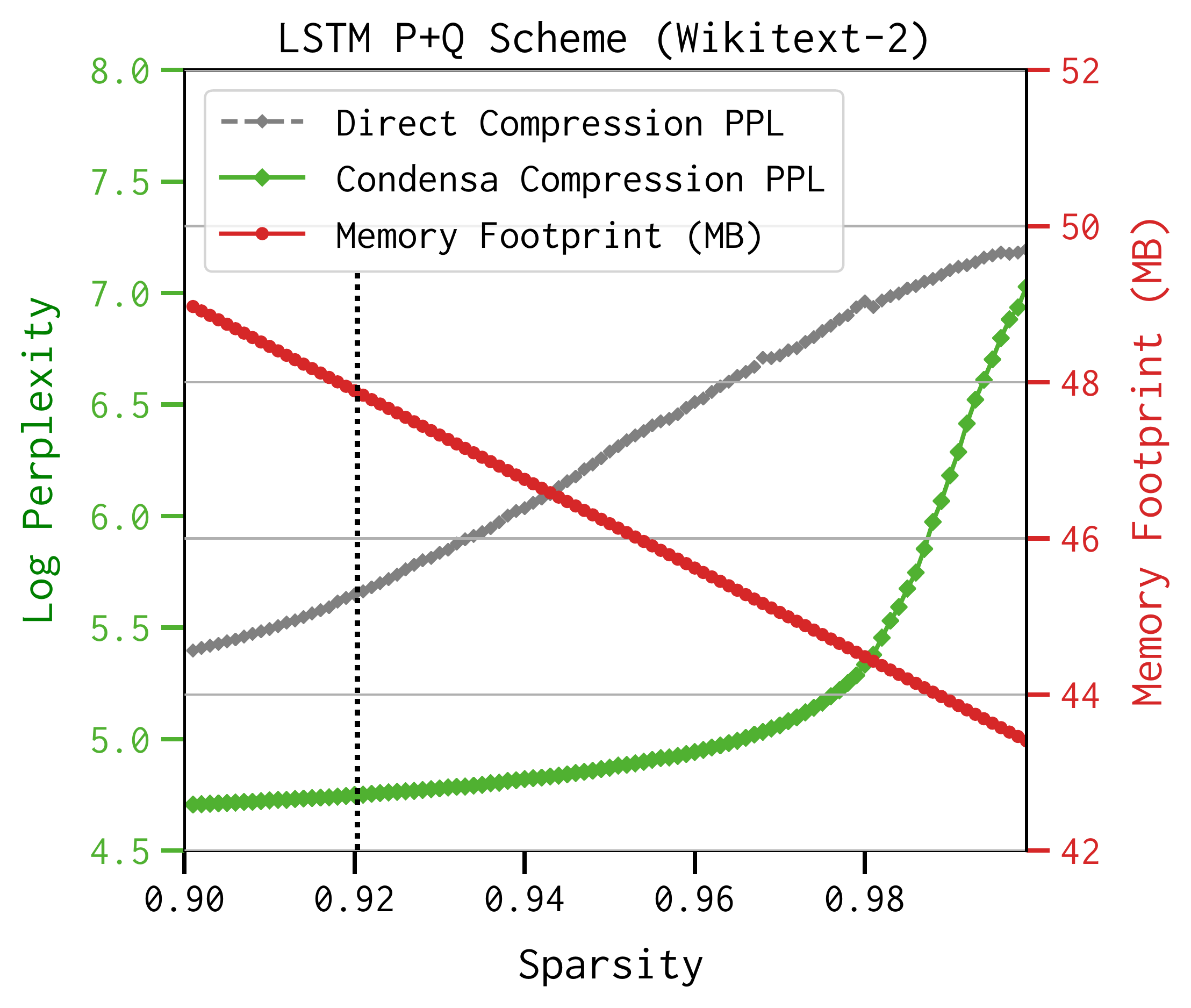}
\includegraphics[height=0.4\linewidth] {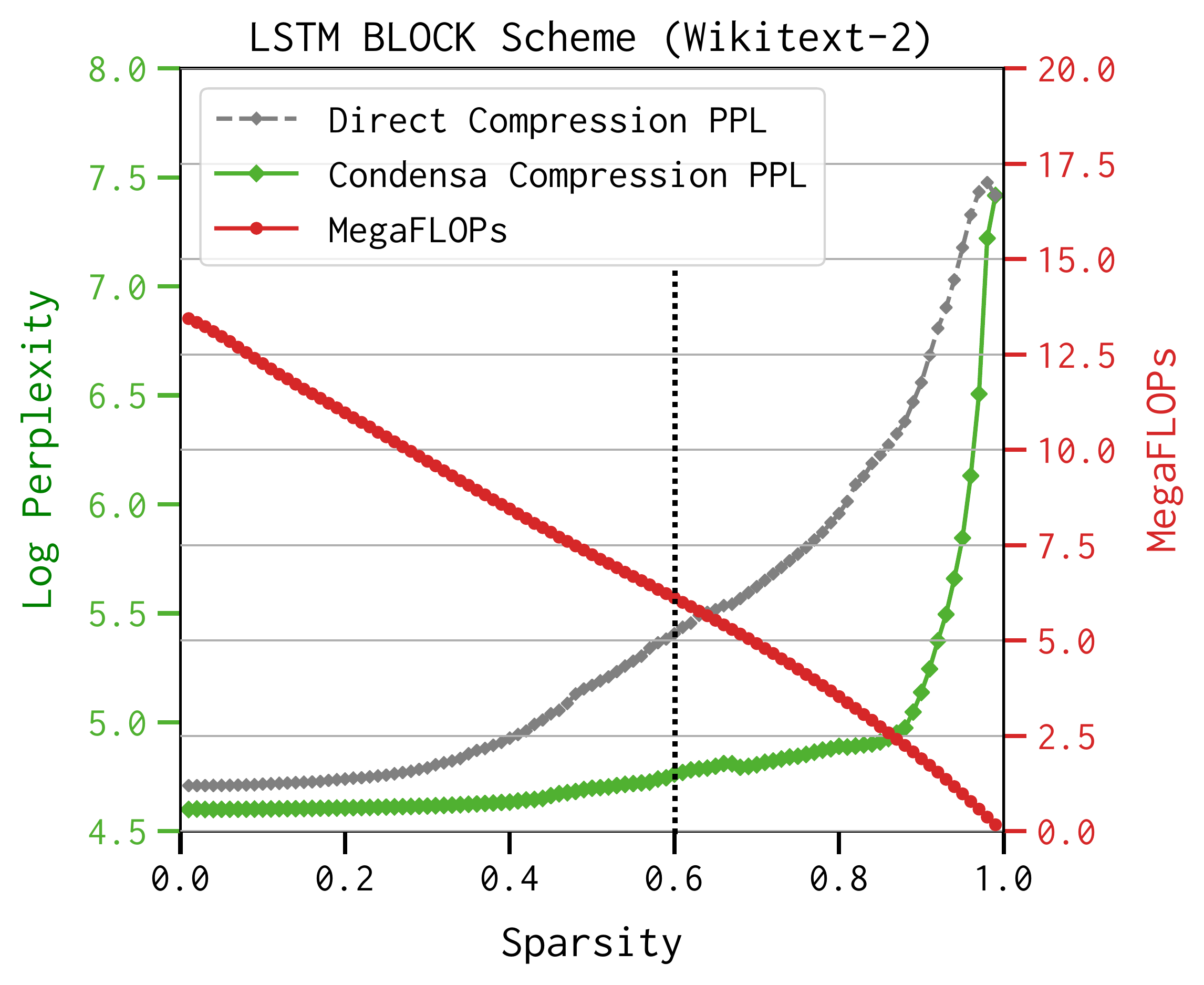}
\caption{WikiText-2 two-layer LSTM results for pruning + quantization (left) and block pruning with block size of 5 (right). Note that the x-axis ranges are different: the plot on the left has sparsity values ranging from $0.9$ to $1.0$ while the one on the right has values ranging from $0$ to $1$.}
\label{fig:profile-lstm}
\end{figure*}

\noindent \textbf{(3) Language Modeling on WikiText-2} 
We trained a 2-layer LSTM model to perform a language modeling task on the WikiText-2 dataset~\cite{merity2016pointer}. We used a hidden state size of $650$ and included a dropout layer between the two RNN layers with a dropout probability of $0.5$. The LSTM received word embeddings of size $650$. For training, we used truncated Backpropagation Through Time (truncated BPTT) with a sequence length of $50$. The
training batch size was set to $30$, and models were optimized using SGD with a learning rate of $20$. This setup is similar to the one used by Yu et al.~\cite{yu2019playing}.

We optimize the networks in each task for two distinct objectives described below:

\noindent \textbf{Objective 1: Minimize Memory Footprint} The memory footprint of a model is defined as the number of bytes consumed by the model's {\em non-zero} parameters. Reducing the footprint below a threshold value is desirable, especially for memory-constrained devices such as mobile phones, and can be accomplished through either pruning or quantization, or both.
For reducing footprint, we define a compression scheme that performs unstructured pruning of each learnable layer (except batch normalization layers), and then quantizes it to half-precision floating-point, yielding an additional 2x reduction.
We denote this scheme by P+Q and implement it using the \algoName library as follows (see Table~\ref{tab:schemes} for the full list of schemes):

\begin{lstlisting}[numbers=none]
from schemes import Compose, Prune, Quantize
scheme = Compose([Prune(), Quantize(float16)])
\end{lstlisting}

\noindent \textbf{Objective 2: Maximize Throughput}
Inference throughput is defined as the number of input samples processed by a model per second, and is commonly used for measuring real-world performance. For CIFAR-10 and ImageNet, we measure hardware inference throughput of the compressed model in the objective function. We use an NVIDIA Titan V GPU with the TensorRT 5 framework to obtain throughput data. For WikiText-2, due to the lack of optimized block-sparse kernels for PyTorch, we measure the floating-point operations (FLOPs) of the compressed model instead as a proxy for inference performance. To improve throughput, we focus on removing entire blocks of non-zeros, such as convolutional filters, since they have been proven to improve performance on real-world hardware~\cite{he2018progressive,gray:2017}. For CIFAR-10 and ImageNet, we use filter pruning, since all the networks we consider are CNNs. In WikiText-2, we employ block pruning with a block size of 5. 

\noindent \textbf{Bayesian Optimizer Settings}
We use a Gaussian Processes prior with the Matern kernel ($\nu = 2.5$), length scale of $1.0$ and $\alpha$ value of $0.1$ with normalization of the predictions. For the GP regressor, the noise level in the covariance matrix is governed by another parameter, which we set to a very low value of $10^{-6}$. For the DR-UCB acquisition function, we use a $\gamma$ value of $0.95$ for all our experiments with a bias towards sampling more in the area of level set, with the intention that the Bayesian optimizer results in a favorable sparsity level in as few samples as possible.
We implemented DR-UCB using the \texttt{fmfn/BO} package~\cite{fmfnbo}.

\noindent \textbf{L-C Optimizer Settings}
The L-C optimizer was configured as follows: for all experiments, we use $\mu_j = \mu_oa^j$, with $\mu_0 = 10^{-3}$ and $a = 1.1$ where $j$ is the L-C iteration. For CIFAR-10 and ImageNet, we use the SGD optimizer in the learning (L) step with a momentum value of 0.9, with the learning rate decayed from $0.1$ to $10^{-5}$ over each mini-batch iteration. We use the Adam optimizer in the L-step of WikiText-2 with a fixed learning rate of $10^{-4}$. We ran between 4000-5000 mini-batch iterations in each L-step, with a higher number of iterations in the first L-step ($30k$ for CIFAR-10 and ImageNet, and $7k$ for WikiText-2) as recommended by~\cite{carreira2018learning}. We ran 5, 30, and 50 L-C iterations for WikiText-2, ImageNet, and CIFAR-10, respectively; compared to CIFAR-10, we ran relatively fewer iterations for ImageNet due to its significantly higher computational cost, and ran an extra 5 fine-tuning iterations instead. We use the same mini-batch sizes as during training for all experiments, and use validation datasets to select the best model during compression (we perform a 9:1 training:validation split for CIFAR-10 since it doesn't include a validation dataset).

\subsection{Results}
\label{sec:results}

We present the memory footprint reductions and inference throughput improvements obtained by \algoName for each of the three tasks we evaluate in Table~\ref{tab:results}. For each task, we list the sparsity obtained by the \algoName Bayesian optimizer ($s^{*}$ in the table), its corresponding accuracy/perplexity (top-1 accuracy, top-5 accuracy, and log perplexity for CIFAR-10, ImageNet, and WikiText-2, respectively), memory footprint reductions using pruning and quantization (column labeled $r_c$), and inference throughput/FLOP improvements using filter/block pruning.
%
We also compare our approach with recent work on automated model compression. For CIFAR-10 and ImageNet, we compare our results with AMC~\cite{he2018amc} and AutoSlim~\cite{liu2019autoslim}, and for WikiText-2, we compare with~\cite{yu2019playing}.
Since AMC~\cite{he2018amc} and~\cite{yu2019playing} do not report actual runtime numbers on hardware, we report the corresponding FLOP improvements instead (values marked $s_F$). We also use FLOP reduction as a metric for LSTM block pruning, as described above.
%

Using the P+Q scheme designed to minimize memory footprint, \algoName is able to obtain compression ratios up to $188\times$, which surpasses those of frameworks such as AutoCompress. While AMC and AutoCompress only report theoretical FLOP improvements on CIFAR-10 and ImageNet, the filter pruning strategy implemented using \algoName yields real-world runtime improvements of up to $2.59\times$ on an NVIDIA Titan V GPU.
Since AMC and AutoCompress do not report the number of samples evaluated to arrive at their solutions, we are unable to directly compare sample efficiencies with these frameworks; however, we notice that \algoName obtains desirable model sparsities using a fixed 10 iterations per search in all experiments.
Finally, while we set the level set to be $2\%$ below the accuracy of the reference model in all our experiments, we notice that \algoName-compressed models often exceed baseline accuracy.

\subsection{Sparsity Profile Analysis}
\label{sec:profile}

Figures~\ref{fig:profile-cifar} and~\ref{fig:profile-lstm} illustrate how a compressed model's accuracy, inference performance, and memory footprint vary w.r.t. sparsities for the CIFAR-10 and WikiText-2 tasks. All three of these functions are {\em assumed to be unknown} in our problem formulation, but we compute them explicitly here to better understand the quality of solutions produced by \algoName. For each figure, compression accuracies (shown in green) are obtained by running the L-C algorithm to convergence for $100$ sparsity values ranging from $0.9$ to $1.0$ (for P+Q), and from $0$ to $1$ for the filter and block pruning schemes; collecting each such point requires between $30$ minutes to $8$ hours of time on a single NVIDIA Tesla V100 GPU. We are unable to show the full profile for ImageNet due to its significantly higher computation cost: collecting each data point for compression accuracy requires over 12 hours of compute time on a node with 8 Tesla V100 GPUs. Inference throughput, FLOPs, and memory footprint data is collected for each compressed model and depicted by red lines in the figures (right-hand-side y-axis). We also show direct compression (DC) accuracies in gray for comparison (DC is described in more detail in Section~\ref{sec:lc}). In each figure, the sparsity found by \algoName is shown as a black vertical dashed line.

We notice three important trends in Figures~\ref{fig:profile-cifar} and~\ref{fig:profile-lstm}: (1) \algoName consistently finds solutions near the `knee` of the L-C accuracy curves, signifying the effectiveness of the DR-UCB acquisition function; (2) local minima/maxima is avoided while optimizing the objective function, demonstrating that the UCB acquisition function for objective function optimization is working as expected, and (3) the knee of the D-C accuracy curves occur at significantly lower sparsity ratios; the L-C optimizer, on the other hand is able to recover accuracies up to much higher sparsities.

\subsection{Layerwise Runtime Performance}

\begin{figure}[ht]
\centering
\includegraphics[width=\linewidth]{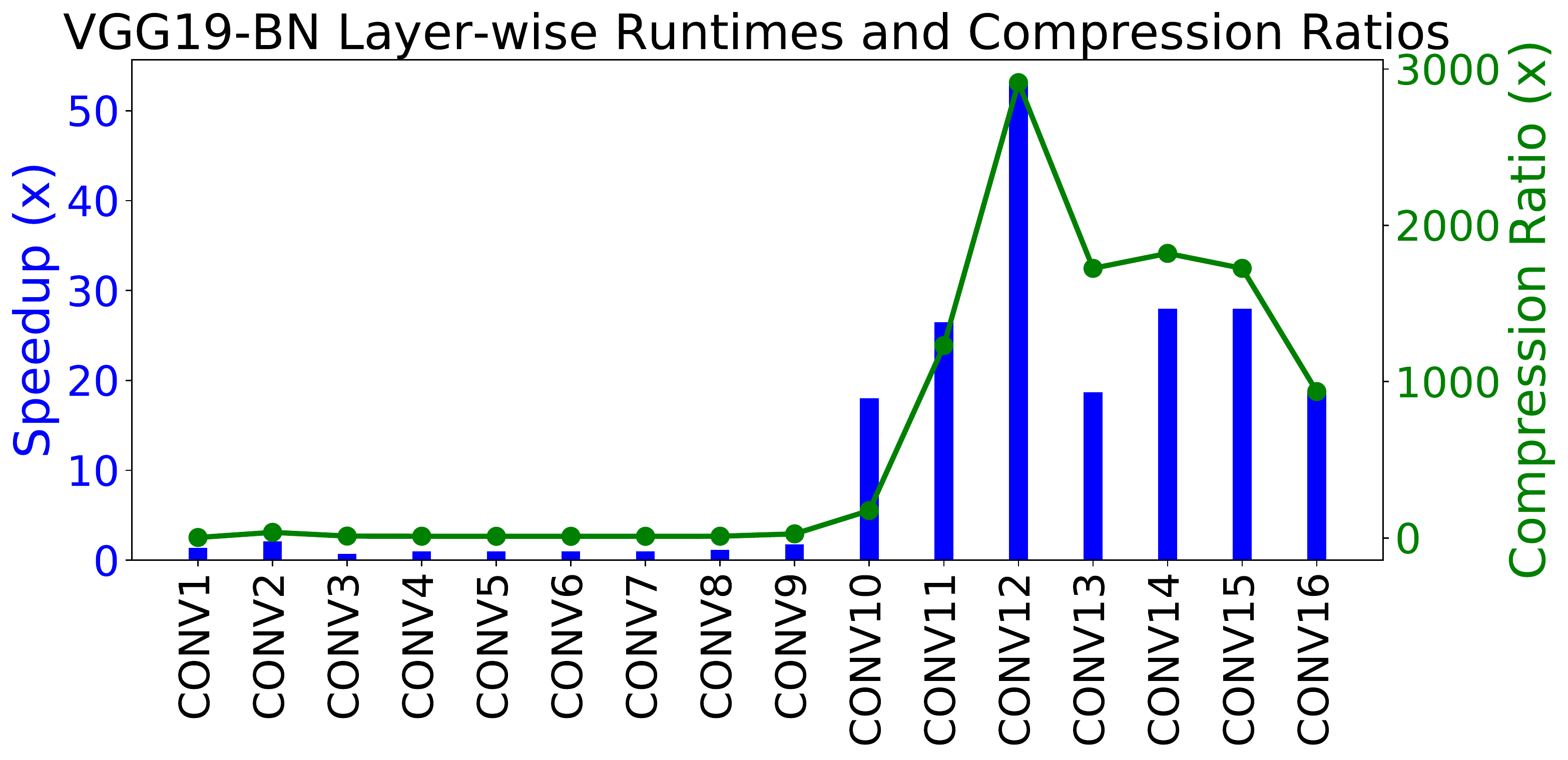}
\caption{TensorRT runtimes and compression ratios of convolutional layers in VGG19-BN (filter pruning).}
\label{fig:layer-speeds}
\end{figure}



In this section, we analyze how improving throughput using compression translates to execution time improvements for each layer on actual hardware. For this experiment, we focus on VGG-19 on CIFAR-10, since it has a relatively simple structure and is easy to analyze on a layer-by-layer basis. We use filter pruning with a target sparsity of $0.79$ (found by the Bayesian optimizer, as shown in Table~\ref{tab:results}) for this experiment.
Figure \ref{fig:layer-speeds} shows layer-by-layer mean runtimes collected over 100 runs using TensorRT (blue bars, left y-axis), and compression ratios (green line, right y-axis) for filter pruning. We only show data for convolutional layers as they dominate computation time for this network.
We make two key observations: (1) runtime speedups on real hardware are largely correlated with compression ratios, but may be affected by hardware and implementation details (e.g., compare \texttt{conv13} with \texttt{conv14} in the Figure), and (2) higher compression ratios and corresponding speedups for the later layers of the network, which indicates that distributing a given global sparsity evenly across network layers may not always be optimal, and algorithms such as L-C are essential to automatically finding desirable distributions of sparsity across layers.

%% file: tex/conclusions.tex
\section{Acknowledgments}
This material is based upon work supported by DARPA under Contract No. HR0011-18-3-0007, NSF award CCF-1704715 and a CIFAR AI Chair award. Any opinions, findings and conclusions or recommendations expressed in this material are those of the author(s) and do not necessarily reflect the views of the U.S. Government. 
Distribution Statement "A" (Approved for Public Release, Distribution Unlimited). 

\section{Conclusions}

This paper has presented \algoName, which is a flexible programming system for model compression and corresponding hyper-parameter optimization.
We have demonstrated \algoName's effectiveness and ease-of-use on a range of state-of-the-art DNNs for image classification and language modeling, and achieved memory footprint reductions of up to $188\times$ and runtime throughput improvements of up to $2.59\times$ using at most $10$ samples per search.
With the initial framework in place, we envision a number of directions to expand on \algoName's capability. For example, we plan to augment automatic sparsity inference with support for additional compression hyperparameters such as block sizes in block-sparsification~\cite{gray:2017}, and data types for quantization. Our long-term goal is a framework that makes model compression easier, more flexible, and accessible to a wide range of users.